# Is ChatGPT like a Nine-year-old child in Theory of Mind? Evidence from Chinese writing

Siyi Cao[ab] [*], Yizhong Xu[d*], Tongquan Zhou[**], Siruo Zhou[c],

a. School of Foreign Languages, Southeast University, Nanjing, China, 211189

b. Department of Chinese and Bilingual Studies, The Hong Kong Polytechnic University, Hong Kong, China

c. School of Foreign Studies, Nanjing University of Posts and Telecommunications, Nanjing, China 210023

d. College of Foreign Languages, Nanjing University of Aeronautics and Astronautics, Nanjing, China, 210016

* Equal contribution

**Corresponding authors: zhoutongquan@126.com

**Abstract**

ChatGPT has been demonstrated to possess significant capabilities in generating intricate human-like text, and recent studies have established that its performance in theory of mind (ToM) tasks is just comparable to a nine-year-old child's. However, it remains unknown whether ChatGPT surpasses nine-year-old children in Chinese writing, a task credibly related to ToM. To justify the claim, this study compared ChatGPT with nine-year-old children in making Chinese compositions (i.e., science-themed and nature-themed narratives), aiming to unveil the relative advantages and disadvantages by human writers and ChatGPT in Chinese writing.

Based on the evaluative framework comprising of four indices (i.e., fluency, accuracy, complexity, and cohesion) to test writing quality, this study added an often-overlooked index "emotion" to extend the framework. Afterward, we collected

120 writing samples produced by ChatGPT and children and used the confirmatory factor analysis (CFA) and structural equation modeling (SEM) to analyze their data for comparison. The results revealed that nine-year-old children excelled at ChatGPT in fluency and cohesion while ChatGPT manifested a superior performance to the children in accuracy. With respect to complexity, the children exhibited better skills in science-themed writing, but ChatGPT better in nature-themed writing. Most importantly, this study unlocked the pioneering discovery that nine-year-old children display more potent emotional expressions than ChatGPT in Chinese writing, providing an instance of evidence that ChatGPT is really even poorer than a nine-year-old child in ToM to some extent.

## 1. Introduction

Artificial intelligence (AI) has shown impressive growth and diversity in recent years, with particular strides being made in the field of natural language learning (Dergaa et al., 2023). Among the vanguards in this arena, OpenAI's ChatGPT stands out with its compelling capabilities to generate intricate, human-like text (De Angelis et al., 2023). This has ignited a fascinating debate about the scope and boundaries of AI's language proficiency (Zhou et al., 2023). Notably, research has revealed that Chinese writing is not ChatGPT's strong suit, but its performance on theory of mind (ToM) tasks is comparable to that of a nine-year-old child (Kosinski, 2023). ToM is the ability to understand that others have perspectives and feelings that are different from one's own (Doherty, 2008). This ability is particularly relevant to writing, in which conveying and interpreting emotions plays a crucial role, for writers wants to convey their perspectives and feelings when interacting with readers. Within this fascinating context, an interesting question arises: is ChatGPT's ToM like a nine-year-old child in Chinese writing?

Prior writing research identifies fluency, accuracy, complexity, and cohesion as key indices to test the quality of written Chinese (Tong et al., 2014; Yan et al., 2012). However, previous studies ignore a crucial component of text quality, that is, "emotion" which is often manifested via expressive language and fosters deep connections with readers (Bohn-Gettler & Rapp, 2014). In order to assess the quality of Chinese compositions more comprehensively, this study introduces an often-overlooked aspect "emotion" as an additional index to enrich the evaluative framework for Chinese writing proficiency. As a result, our analysis is to encompass five pivotal indices, including fluency, accuracy, complexity, cohesion, and emotion. Each index representing one dimension is to undergo a detailed assessment with resort to specific variables, as is to be elaborated in the subsequent part. Our objective is to shed light on the areas where ChatGPT excels or falls short in the five indices compared to human writers, particularly nine-year-olds, in the context of Chinese writing.

### 1.1 ChatGPT's development and language capability

ChatGPT is premised on the transformative GPT-4 model, a deep learning architecture designed for understanding and generating natural language. The model's training involves two core steps: a generative, unsupervised pre-training stage (which exposes the model to large quantities of unlabeled data) followed by a discriminative, supervised fine-tuning stage (which optimizes the model for specific tasks) (Chung et al., 2021). This training approach has equipped ChatGPT with the proficiency to perform a variety of tasks such as automated summarization (Ray, 2023), machine translation (Peng et al., 2023), and question-answering (Tan et al., 2023).

ChatGPT demonstrates impressive conversational abilities due to its training process and architecture (Transformer & Zhavoronkov, 2022). Trained on a vast dataset of online texts spanning diverse topics and genres, ChatGPT can comprehend queries and generate coherent, human-like responses (Ray, 2023). Furthermore, reinforcement learning techniques enabled the model to refine its skills through human feedback (Liu et al., 2023). This allows ChatGPT to understand user intent, adjust its responses accordingly, and sustain context-aware dialogues. Through this specialized training regime, ChatGPT has become adept at natural conversation, showcasing advanced language understanding and generation capabilities (Vaishya et al., 2023).

While ChatGPT has garnered attention for its impressive capabilities and practical applications, it has also stirred debates in the linguistics sphere. Distinguished linguists including Noam Chomsky expressed concerns that ChatGPT's approach devalues humans' innate *linguistic competence*. According to Chomsky, humans intuitively understand and form grammatically accurate sentences (Collins, 2007), suggesting that our language competence is a fundamental cognitive function, not mere imitation (McGilvray, 2014). Chomsky's criticism stems from ChatGPT's role as a data duplicator, a stance at odds with his viewpoint on human language (Chomsky et al., 2023). Similarly, Steven Pinker advocated '*language instinct*,' asserting that our innate, genetically coded language learning ability distinguishes humans from other species (Pinker, 2003). Chinese scholar Yuan favored this point, arguing that ChatGPT was ill-suited as a technological interface for human machine interaction. He believed that effective language usage demands embodied intelligence, incorporating lexical grounding and environmental affordances in embodied cognition (Yuan, 2023). In light of these views, ChatGPT, though able to replicate data patterns, lacks the genuine linguistic intelligence required for language acquisition and use.

Contrasting with prior critiques, recent studies highlighted ChatGPT's exceptional language capabilities. Jiao et al. (2023) found that ChatGPT's translation skills could compete with established commercial products like Google, DeepL, and Tencent, affirming its prowess as a translator. In terms of discourse skills, Benzon (2023) identified that ChatGPT could engage in analogical reasoning, interpret films and stories, and comprehend abstract concepts like justice and charity. Additionally, it could modulate its discourse level to match children of different ages. With regard to language usage, Cai et al. (2023) assessed 12 tasks involving sound, words, syntax, meaning, discourse, demonstrating that ChatGPT's choices paralleled human decisions in 10 tasks, though it disfavor shorter words for less informative content and lacked context utilization to clear syntactic ambiguities. Regarding psychological-linguistic skills, Kosinski (2023) confirmed that ChatGPT's performance was just akin to a nine-year-old child's on theory of mind (ToM) tasks. For instance, given a narrative involving the relocation of a cat, ChatGPT accurately deduced the cat's location. Remarkably, ChatGPT achieved 100% accuracy across all 20 tasks, showcasing advanced cognitive abilities (Kosinski, 2023). These findings pose the intriguing question of whether ChatGPT's writing capability might match those of nine-year-old children.

## 1.2 Chinese writing by ChatGPT and nine-year-old children

Writing, deemed a pivotal skill in our everyday lives (Nasser, 2016), has undergone a significant transformation with the introduction of AI authors like ChatGPT last year. Previously, the writing domain was solely occupied by human writers; this landscape, however, dramatically shifted with the advent of AI. As highlighted in numerous academic works, ChatGPT promises to deliver an array of benefits such as language support, translation, editing, and proofreading (Shahriar & Hayawi, 2023). In practice, ChatGPT can augment researchers' linguistic proficiency and writing abilities by providing immediate feedback on grammar, syntax, spelling, and vocabulary, consequently enhancing the overall standard of their manuscripts (Zhou et al., 2023). Additionally, ChatGPT can rapidly produce text on intricate or technical subjects. However, it's important to remember Chomsky's observation that this could potentially be characterized as high-tech plagiarism (Chomsky et al., 2023).

The proficiency of ChatGPT in Chinese writing has been conjectured to fall substantially short of its capabilities in high-resource European languages such as English and German, given the status of Chinese as a low-resource or distant language[1] (Jiao et al., 2023). A recent study, seemingly validating this assumption, challenged ChatGPT to craft a fictional Chinese composition based on the prompt, "请依照题目完成以下的作文'你曾经犯下一个错误，伤透了父母的心。写出犯错经过和你的后悔之情。(*You once made a mistake that deeply hurt your parents. Write about the process, your feelings of regret, and the lessons learned*)" (Rudolph et al., 2023). Although ChatGPT created a response consistent with the topic, the composition was found to lack structure and was riddled with grammatical errors, suggesting a deficiency in its handling of Chinese language. Nevertheless, the study just relied on one sample and lacked statistical analysis of data, so the true proficiency of ChatGPT in Chinese writing remains an open question.

Chinese writing is generally considered an important and difficult skill for nine-year-old children (third grader in China), for they just start to learn how to write in this intricate language and possess limited life experiences (Wang, 2021). At this developmental stage, their writing is generally characterized by simple narrative constructs, leading to the compositions that may lack depth and complexity (Zhang, 2023). Along with the practice, the pupils frequently make a multitude of errors in their compositions, including lexical and syntactic mistakes (Sun, 2023). Moreover, the increased occurrence of homophonic and similar-shaped characters contributes to typographical errors, making accurate recognition of Chinese characters more challenging (Wang, 2023). They also commonly make mechanical errors, such as misplacing periods or commas (Yan et al., 2012). As mentioned in 1.1, the advanced cognitive capabilities of ChatGPT, including the theory of mind (ToM), parallel those of a nine-year-old child. This fact leads us to raise an interesting question: could ChatGPT outperform the pupils' Chinese writing, a task requiring advanced cognitive abilities?

---

[1] High-resource languages refer to the languages such as English, French, German, and Spanish, characterized by extensive linguistic data and research tools available for academic and computational use. Low-resource languages refer to the languages with limited linguistic and computational resources, making advanced processing and analysis challenging (Jiao et al., 2023).

Narrative writing, characteristic of vivid depiction of events, experiences, or emotions, demands a creative approach towards elements like characters, plot, setting, and dialogue. This type of writing provides an avenue for individual expression and entertainment (Oatley, 1995). The writing skills and strategies of narrative enable authors to demonstrate their cognitive capabilities as they probe into character motivations and perspectives. The writing also necessitates thoughtful consideration and logical reasoning, with special attention to aspects like plot progression (Prado et al., 2015). Furthermore, a high level of language proficiency and an aptitude for descriptive, captivating language are essential components of narrative writing (McFadden & Gillam, 1996). When compared with other intricate writing like argumentation, narrative seems to be a more accessible and engaging way for nine-year-olds to display their critical thinking and language abilities (Xu, 2018). Against this background, narrative provides a sensible benchmark to evaluate and gauge the writing competency of both the children and ChatGPT.

**1.3  Assessment of Chinese writing**

Thus far, few studies have established a scoring method for assessing the compositions by young Chinese children. Building upon the English scoring system set by Wagner et al. (2011), Yan et al. (2012) devised a scoring framework specifically for 9-year-old Chinese children in Hong Kong. This framework is segmented into two main dimensions, content and organization. The content dimension evaluates relevance, breadth, and depth, while organization assesses sentence and paragraph structures, key elements, and intelligibility. Inspired by Yan et al. (2012), Tong et al. (2014) introduced a modified scoring model for 11-year-olds in Beijing, establishing the criteria for grammatical errors and punctuation accuracy. Yet, the existing assessment approaches have not sufficiently addressed a crucial component of narrative writing, "emotional intensity", which is often manifested via expressive language and plays a vital role in enriching the narrative, enabling writers to connect more deeply with their readers (Bohn-Gettler & Rapp, 2014). Given this gap, this study attempts to propose a more comprehensive scoring approach that preserves the established scoring criteria while integrating an assessment of emotional resonance.

As mentioned above, the scoring method in the current study spans five indices: fluency, accuracy, complexity, cohesion, and emotion (Table 1). The first index "fluency" is measured by total number of sentences and T-units, for there is a strong correlation between number of sentences and writing fluency (Cahyono et al., 2016). A T-unit is characterized as an independent clause accompanied by any subordinate elements linked to it, whether they be phrases or clauses (Wagner et al., 2011). Jiang (2013) validated the T-unit as a trustworthy tool for scrutinizing the progression of Chinese writing.

The second index "accuracy" is tested by calculating spelling errors, grammatical errors, and punctuation errors, as posited by Tong et al. (2014). Here, grammatical errors refer to those like disordered word sequences, fragmented sentences, misnomers, collocation mistakes, redundancy, and ambiguous or incorrect usage.

The third index "complexity" is investigated by three metrics, the frequency of uncommon words, the number of idioms and the count of unrepeated words. The prevalence of uncommon words and idioms directly correlates with an article's complexity, that is, a higher frequency tends to render articles more challenging to comprehend (Nippold, 2000). Likewise, the number of unrepeated words in a text reflects vocabulary diversity and more unrepeated words points to higher lexical complexity, signaling a broader vocabulary range (Isaacson,1988).

The fourth index "cohesion" is measured by two factors, the total number of connectives and the number of accurately used connectives. As a common practice, the use of connectives is integral to the cohesion of a written piece, so their numbers count much in the evaluation of writing quality (Crossley & McNamara, 2012).

The fifth index "emotion" is tested via the intensity of the emotion. Since machines currently lack the depth of human empathy and emotion, their writings might not exhibit strong emotional intensity Carlbring et al. (2023). Consequently, this may serve as a distinguishing factor between human and AI-generated writing.

Table 1. The description of five indices

| Dimensions | Indicators | Abbreviation |
|---|---|---|
| Fluency | total number of sentences | TNOS |
|  | T-units | TUNIT |
| Accuracy | the number of spelling errors | NOSE |
|  | the number of grammatical errors | NOGE |
|  | the number of punctuation errors | NOPE |
| Complexity | the number of uncommon words | NOUCW |
|  | the number of idioms | NUI |
|  | the number of unrepeated words | NOURW |
| Cohesion | the number of correct connectives | NOCC |
|  | total number of connectives | TNOC |
| Emotion | the intensity of emotion | IOE |

In summary, ChatGPT has to date achieved two major milestones in generating linguistic text. The first reveals that ChatGPT is comparable to a nine-year-old child in ToM. The second demonstrates that ChatGPT exhibits exceptional linguistic capacities (e.g., text generation and language translation) even far beyond human adults. Along with its remarkable advancements are ChatGPT's some issues to solve. To start with, ChatGPT mainly relies on the sources of English and its data size is relatively small in other languages like Chinese, presumably exposing its deficiency in handling Chinese texts. In addition, even though ChatGPT is on par with a nine-year-old child on ToM tasks, it is uncertain whether ChatGPT can surpass nine-year-olds in Chinese writing. These two aspects motivated our study as an attempt to answer two research questions relevant to Chinese writing:

(a) Does ChatGPT outperform nine-year-old children in Chinese narrative writing?

(b) How do ChatGPT and the children perform in the five indices used to assess writing (fluency, accuracy, complexity, cohesion, and emotion)?

## 2. Method

### 2.1 Participants

30 nine-year-old children (third-grader) from a primary school in the southern China were recruited to participate in this study on a voluntary basis. The children were composed of 14 females and 16 male with the average age between 8 and 10 ($M = 9.04$, $SD = 0.73$). Prior to the experiment, each participant read and signed the informed consent on a written paper. The experiment received approval from the Human Research Ethics Committee of the university the first author is affiliated with.

### 2.2 Procedure

This study was conducted in a controlled classroom, in which the 30 nine-year-old participants were instructed to compose two Chinese narrative essays based on given topics, resulting in 60 articles in total. All the participants were required to finish the task within two natural-class time (around 80 minutes) with a short interval (about 10 minutes) between classes.

The first task required participants to write about science and everyday life. Specifically, they wrote according to the Direction I as below.

Direction I: "科学离不开生活，科技的进步来源于人类对美好生活的想象创造。未来的你如果是一名科技工作者，你想发明什么？它是什么样子的？有哪些功能？让我们把它写出来吧！请完成习作，300字左右。"

(English version: *Science is inseparable from life, and the advancement of technology stems from humanity's imagination and creation of a better life. If you become a technologist in the future, what would you like to invent? What does it look like? What functions does it have? Let's write such a composition around 300 words.*)

The second task switched the topic into nature and its reflection, wherein the students were requested to write according to the Direction II as below.

Direction II: "这一单元，我们看到了作者笔下一幅幅动人的画面，那乡下的动物、植物、人家，那孩子眼中小小的神奇的天窗……这些画面中都饱含着作者的情感。请以《我心上的这幅画》为题，写一篇习作，表达自己的真情实感。300字左右。"

*(English version: In this unit, we witnessed a series of touching scenes penned by the author: the animals, plants, and households of the countryside, and the little magical skylight in a child's eyes... Each of these images is imbued with the author's emotions. Please write a composition titled "The Painting in My Heart" around 300 words, expressing your genuine feelings.)*

For the sake of comparison, ChatGPT was employed to produce 60 unique essays (30 for each topic) using the same topics as adopted by the children, ensuring no repetition in content.

## 2.3 Coding

A total of 120 Chinese compositions (30×2×2) were collected from the 30 children and ChatGPT under science-themed and nature-themed directions as detailed above. Since the compositions produced by children were all handwritten, they were digitized for statistical analysis. To guarantee consistency between the handwritten and digital versions, another experimental assistant verified the congruence of both forms and added the small ignored details, such as spelling errors.

To construct an assessment model encompassing the five indices, we meticulously collected the data for each indicator within every index. To start with, the indices including total number of sentences (TNOS), number of spelling errors (NOSE), number of punctuation errors (NOPE), number of uncommon words (NOUCW), number of idioms (NUI) were extracted according to the website *https://xiezuocat.com/*, which serves as a valuable resource for writing assessment, offering scores across various indices to enhance writing quality. Afterward, two master students of English majors were invited as independent raters to assess the website's outputs based on the numbers. Additionally, intensity of Emotion (IOE) was extracted using SnowNLP library in Python, which is one of the most popular Python libraries for sentiment analysis for Chinese language natural language processing (Yu et al., 2021).

Apart from the above, the indicators like TUNIT, NOGE, NOURW, number of correct connectives, and total number of connectives were rated and graded by another two assistants. Inter-rater reliabilities for these elements were calculated on the basis of 120 samples and manifested as Pearson's correlation coefficients, ranging between .76 and .89. The final score was determined by averaging the scores from both raters.

## 2.4 Data analysis

Firstly, a Confirmatory Factor Analysis (CFA) was adopted to validate a predefined assessment model for the compositions across five dimensions, using the

"lavaan" package in R (R Core Team, 2016). This was followed by testing the data's fit to the model so as to affirm its validity. Specifically, the CFA sought to determine if the 11 linguistic factors could be effectively predicted by five measured variables. The goodness-of-fit for the specified model was gauged using multiple statistical tests and indices. The chi-square test, for instance, was employed to assess the null hypothesis that the covariance matrix from our model aligns with the observed data covariance matrix. Ideally, this test was expected to produce a non-significant result ($p > .05$) to uphold the null hypothesis. Given our relatively small sample size ($N = 120$), we selected the CFI and SRMR to judge the model's fit, for the RMSEA and TLI can be overly critical in smaller samples. With the predetermined criteria, we considered models exhibiting a CFI $\geq .90$ and SRMR $\leq .07$ to have a satisfactory fit.

Secondly, a structural equation model (SEM), also accessed by the "lavaan" package in R, was applied to build a model in order to showcase the relationship between "group" (nine-year-old children vs. ChatGPT), "writing type" (science vs. nature) and five latent measured factors: fluency, accuracy, complexity, cohesion and emotion. Given the limited sample size ($N = 120$), we relied on the CFI and SRMR to evaluate model fit, for both RMSEA and TLI tend to inaccurately reject models when the sample size is small. Based on our predetermined criteria, a model was considered to have an adequate fit if the CFI was $\geq .90$ and the SRMR was $\leq .07$.

Thirdly, a two-way analyses of variances (ANOVAs) was used to discern differences across the five linguistic dimensions in light of "group" and "writing type", which was achieved by resort to the EMMEANS function from the bruceR package (Bao, 2023). Whenever a significant main effect emerged, we performed multiple comparisons using the Tukey method.

## 3. Results

The CFA modeled five latent linguistic dimensions for a total of 11 factors, suggesting that each linguistic dimension (i.e., fluency, accuracy, complexity, cohesion and emotion) could be predicted by various separate linguistic variables. The results showed that the model had adequate fit based on a priori criteria ($\chi^2 = 91.6$, $df = 35$, CFI = .93, SRMR = .07) and factor loadings were detailed in Table 2.

Table 2. Factor loadings for model indices

|  | *λ* | *S.E.* | *p* |
|---|---|---|---|
| Fluency | | | |
| TNOS | .49 | .50 | < .001 |
| TUNIT | .93 | .82 | < .001 |
| Accuracy | | | |
| NOSE | .93 | .34 | < .001 |
| NOGE | .56 | .07 | < .001 |
| NOPE | .39 | .22 | < .001 |
| Complexity | | | |
| NOUCW | .93 | 1.7 | < .001 |
| NUI | .76 | 1.29 | < .001 |
| NOURW | .27 | .08 | < .001 |
| Cohesion | | | |
| NOCC | 1.0 | .27 | < .001 |
| TNOC | .99 | .26 | < .001 |
| Emotion | | | |
| IOE | 1.0 | .01 | < .001 |

Additionally, the SEM modeled the variables of "group" (nine-year-old children vs. ChatGPT) and "writing type" (science vs. nature) as predictors of five latent linguistic dimensions. As shown in Fig 1, the model had adequate fit based on set a priori criteria ($\chi^2$ = 146.71, $df$ = 47, CFI = .91, SRMR = .07), indicating that the variables of "group" and "writing type" were significantly associated with five latent linguistic dimensions. Specifically, "group" was significant positively associated with fluency ($B$ = .25, $p$ < .05), accuracy ($B$ = .92, $p$ < 0.001) and emotion ($B$ = .39, $p$ < 0.001). In contrast, "writing type" was significant negatively associated with complexity ($B$ = -.55, $p$ < 0.001), cohesion ($B$ = -.64, $p$ < 0.001) and emotion ($B$ = -.49, $p$ < 0.001).

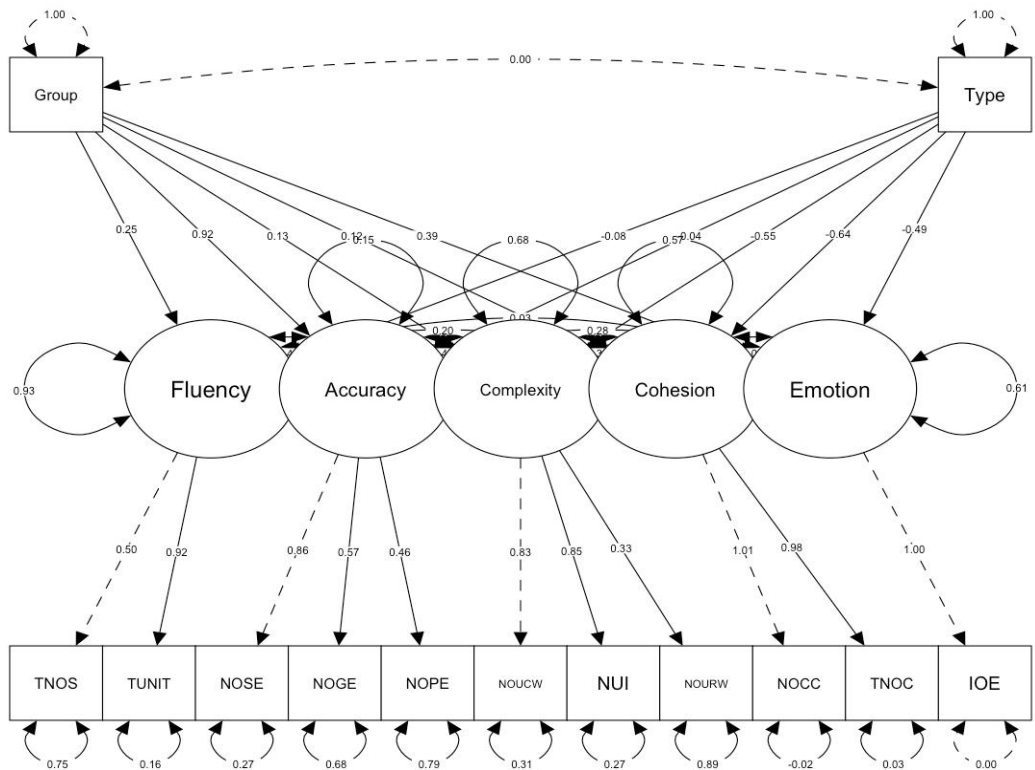

Figure 1. Structural equation model with "Group" and "Type" as predictors of five linguistic factors (fluency, accuracy, complexity, cohesion, and emotion). *Note that* all parameters estimates are standardized. Dashed lines represent the indicators that are fixed in the fixed loading method.

Afterward, a two-way analyses of variance (ANOVAs) was adopted to assess whether ChatGPT outperformed the nine-year-old children on five linguistic factors across different types of Chinese writing (Table 2). The factors examined included fluency, accuracy, complexity, cohesion, and emotion, which served as the dependent variables. The independent variables were "Group" (i.e., nine-year-old children vs. ChatGPT) and "Writing Type" (i.e., science vs. nature).

For fluency, a significant main effect was observed for the "Group" variable ($F(1, 116) = 8.623, p < .01$), but no significant interaction between "Group" and "Writing Type" was found ($F(1, 116) = .78, p = .379$). Pairwise comparisons revealed that the children exhibited higher fluency levels than ChatGPT in science-themed writing ($\beta$(nine-year-old children–ChatGPT) = 1.689, $t(116) = 2.70, p < .01$).

For accuracy, there appeared both a significant main effect for "Group" ($F(1, 116) = 226.15, p < .001$) and a significant interaction between "Group" and "Writing Type" ($F(1, 116) = 8.30, p < .01$). Pairwise comparisons showed that in nature-themed writing, the children demonstrated higher values than ChatGPT ($\beta$(nine-year-old children–ChatGPT) = 3.514, $t(116) = 8.60, p < .001$). Similarly, in science-themed writing, the children also outperformed ChatGPT in values ($\beta$(nine-year-old children–ChatGPT) = 5.17, $t(116) = 12.67, p < .001$). These accuracy values were derived from the counts of three different types of errors and the higher the values, the lower the accuracy. That is to say, these results demonstrated that ChatGPT exhibited greater accuracy than nine-year-old children in both types of writing.

For complexity, we observed two significant main effects: one for the "Group" variable ($F(1, 116) = 4.62, p < .05$) and another for "Writing Type" ($F(1, 116) = 18.59, p < .001$). There appeared a significant interaction between "Group" and "Writing Type" ($F(1, 116) = 5.79, p < .05$) as well. Pairwise comparisons revealed that in science-themed writing, the children displayed greater complexity than ChatGPT ($\beta$(nine-year-old children–ChatGPT) = 14.78, $t(116) = 3.22, p < .01$). Intriguingly, ChatGPT exhibited higher complexity in nature-themed writing relative to its performance in science-themed writing ($\beta$(science–nature) = -21.80, $t(116) = -4.750, p < .001$).

For cohesion, there were two significant main effects of "Group" ($F(1, 116) = 4.10, p < .05$) and "Writing Type" ($F(1, 116) = 92.53, p < .001$). Furthermore, a significant interaction effect between "Group" and "Writing Type" was also observed ($F(1, 116) = 9.17, p < .001$). Pairwise comparisons suggested that in science-themed writing, the children had greater cohesion than ChatGPT ($\beta$(nine-year-old children–ChatGPT) = 2.864, $t(116) = 3.57, p < .001$). Interestingly, both ChatGPT and the children exhibited higher levels of cohesion in nature-themed writing as compared to their performance in science-themed writing (ChatGPT: $\beta$(science–nature) = -7.17, $t(116) = -8.94, p < .001$; nine-year-old children: $\beta$(science–nature) = -3.74, $t(116) = -4.66, p < .001$).

For emotion, two significant main effects were identified, the "Group" variable ($F(1, 116) = 34.82, p < .001$) and the "Writing Type" ($F(1, 116) = 54.62, p < .001$). Additionally, a significant interaction effect between "Group" and "Writing Type" was observed ($F$(1, 116) = 23.26, $p$ < .001). Pairwise comparisons showed that the children displayed greater emotional depth than ChatGPT in both science-themed and nature-themed writing ($β$(nine-year-old children–ChatGPT) = .182, $t$(116) = 7.58, $p$ < .001; $β$(nine-year-old children–ChatGPT) = .22, $t$(116) = 9.40, $p$ < .001). While ChatGPT manifested higher levels of emotion in nature-themed writing in comparison with its performance in science-themed writing ($β$(science–nature) = -2.21, $t$(116) = -8.63, $p$ < .001), the children didn't show significant difference in emotional expression between nature-themed and science-themed writing ($β$(science–nature) = -.04, $t$(116) = -1.82, $p$ = .07).

## 4. Discussion

The current study compared the quality of Chinese writing generated by ChatGPT and nine-year-old Chinese children in terms of five critical indices of language performance: fluency, accuracy, complexity, cohesion, and emotion. The results revealed that the children outperformed ChatGPT in fluency, cohesion, and emotion whereas ChatGPT demonstrated superiority over the children in accuracy. With respect to complexity, the children performed better than ChatGPT in science-themed writing, but the situation reversed in nature-themed writing.

What follows are dedicated to discussing the potential factors that may contribute to these results across the five discourse indices and their correlations in narrative writing.

### 4.1 Fluency: children > ChatGPT

The statistical analysis revealed that the children exhibited greater fluency than ChatGPT, as measured by the TNOS and TUNIT. That is, the children have more TNOS ($p$ < .05) and TUNIT ($p$ < .01). This outcome appears to be attributed to two reasons as below.

Firstly, the length of Chinese compositions produced by ChatGPT frequently fell short of the 300-word benchmark, suggesting a possible misunderstanding of the prompt "*300 字左右*". While "*左右*" is roughly tantamount to "around" in English, ChatGPT might not interpret it accurately according to the given prompt. To validate this hypothesis, the researchers presented several analogous prompts like "*200 字左右*" and "*500 字左右*" to ChatGPT. The resulting text was consistently either significantly shorter than, or exceeded, the 200/500-word guidelines. This fact suggests a potential oversight by ChatGPT's developers regarding the term "*左右*" in Chinese, requiring further research to identify and address such a challenge.

Secondly, data-training limitations of ChatGPT leads to fewer TUNITs than the children. In fact, ChatGPT's responses are shaped by its vast training data, restricting its ability to generate innovative ideas (Abdullah & Jararweh, 2022). As a result, ChatGPT often generates sentences that are directly related to the topic at hand, without venturing into deeper or more imaginative descriptions, as illustrated in (1).

(1) "这幅画中有着美丽的乡村风景和活泼可爱的农村孩子 *(This painting features beautiful countryside scenery and lively, adorable rural children)*".

In contrast, the children are good at making intriguing sentences, reflecting their imagination and personal experiences. For instance, in a nature-themed composition, a child described not only the act of climbing but also the breathtaking view upon reaching the top, as shown in (2).

(2) "我马不停蹄的往观景台那去，看到了比火烧云还美的一景 *(I hastily headed towards the observation deck and saw a scene even more beautiful than the burning clouds)*".

Evidently, children's writing contains a multitude of sub-events under a central theme, whereas ChatGPT's output looked as if more narrowly focused. According to the definition of TUNIT, a TUNIT consists of a main independent clause and its associated dependent clauses. In other words, each TUNIT represents a single, coherent thought or idea, often corresponding to the description of an event. The analysis of writing samples reveals that children generally construct more TUNITs as compared to ChatGPT, indicating that children typically describe more events within their narratives.

According to Brown and Klein (2011), there is a positive correlation between text length (equivalent to TNOS in this study) and ToM in English narrative writing. That is, the nine-year-old children produced longer texts (higher TNOS) than ChatGPT in our test. In Kosinski (2023), ChatGPT exhibits commendable ToM capabilities, evident in its strong empathy and ability to anticipate others' thoughts, thereby paralleling the cognitive abilities of nine-year-olds. The study reveals similar result that ChatGPT is weaker than the children in ToM with regard to text length in Chinese writing.

**4.2 Accuracy: ChatGPT > children**

Our data show that ChatGPT outperformed the nine-year-old children in accuracy, a metric assessed by the number of spelling errors (NOSE), grammatical errors (NOGE), and punctuation errors (NOPE). Subsequent analysis elucidates that ChatGPT commits fewer NOSE ($p < .01$), NOGE ($p < .01$), and NOPE ($p < .01$) in comparison with the children. Such a result is supposed to result from the following three aspects.

Above all, the intrinsic complexity of Chinese orthography significantly contributes to the abundance of spelling errors observed in the the children's writings. Chinese characters operate as logograms, that is, each one symbolizes a word or a meaningful component of a word. In contrast to alphabetic systems, where letters signify sounds, every Chinese character possesses a distinct structure and form. This inherent complexity is a fertile ground for errors as children navigate through the learning process (Sun, 2023). Furthermore, the prevalence of homophones and homographs can induce confusion and result in mistakes for children at this developmental stage (Wang, 2023).

Furthermore, the children still struggle with abstract concepts like grammatical rules. According to the cognitive developmental theory of Piaget (1976), children undergo a series of developmental stages, with each stage marked by unique cognitive capabilities. At around 9 years old, children generally find themselves in the "Concrete Operational Stage" of development, a phase extending roughly from 7 to 11 years. In this stage, children start developing a more organized and rational thought process. They can understand basic grammatical structures and rules as they can think logically about concrete events and situations they encounter. However, their understanding is often tied to tangible and visible situations, concepts, and objects, and they feel hard to capture abstract grammatical rules (Babakr et al., 2019), yielding many grammatical errors as a consequence.

In addition, the children are learning the usage of punctuation and often struggling with utilizing them correctly. As noticed by Tang (2018), the learning of punctuation during the primary school stage is segmented into three phases. In the initial phase, targeting 1st and 2nd graders, the students are introduced to the usage of common symbols (e.g., commas and periods) within texts. The second phase focusing on 3rd and 4th graders, progresses to the comprehension of more intricate punctuation like colons and quotation marks. The final phase shifts the emphasis to the accurate application of these punctuation marks. According to this division, our students are typically in the second phase and so demonstrate unfamiliarity with appropriate punctuation in their writing. This is evident, for instance, in sentences like (3) where punctuation may be used inaccurately. However, ChatGPT has been trained by numerous data and is unlikely to have spelling, grammatical and even punctuation error (Bishop, 2023).

(3) "*每当我们去散步总会看见有人把桌子摆到门口吃 (Whenever we go for a walk, we often see people set up tables at the doorway for eating).*"

### 4.3 Complexity: children > ChatGPT (science-themed writing)

The results revealed that the nine-year-old children adopted more complexity than ChatGPT in science-themed writing, and this situation was reversed in nature-themed writing. Specifically, the children used a great number of complex words, idioms, and unrepeated words than ChatGPT in science-themed compositions, but ChatGPT did better than the children in nature-themed writing. This complexity reversal results from the discrepancy in thematic familiarity, natural inclination and input constraints, to be elaborated below.

Firstly, the increased vocabulary and thematic familiarity that children possess regarding science-themed subjects could be a contributing factor. The selected children in our experiment are undergoing crucial cognitive development and learning stages (Piaget, 1976), easily exposed to a wide range of vocabulary and diverse learning materials, including those on scientific subjects (Lefa, 2014). This exposure likely enhances their utilization of uncommon words, idioms, and unrepeated words in science-themed writing. Recent introductions to new scientific concepts may predispose them to employ a diverse range of words and expressions to articulate scientific phenomena. By comparison, ChatGPT programmed to process a bulk of textual data, doesn't "learn" through or from experience as children do. This performance may result from the shortage of cognitive interaction of ChatGPT, as reported by Xu et al. (2023), for ChatGPT struggles to fully capture motor aspects of conceptual knowledge like actions with foot/leg (Xu et al., 2023). In this way, ChatGPT lacks children's natural curiosity and interpersonal embodiment while producing science-themed content, just relying instead on patterns derived from its training data. This difference could reduce ChatGPT's linguistic diversity in such contexts, as compared to the writing performance of nine-year-olds, who write based on direct cognitive growth and hands-on discovery.

Secondly, the natural inclination of children towards creativity and imagination is noteworthy (Shidiq, 2023). Children's vivid imaginations and inherent creative tendencies drive them to employ a richer, more varied vocabulary, especially in subjects they find intriguing or have newly acquired knowledge from. This intrinsic creativity and imagination may find a more tangible expression in their science-themed compositions, particularly if they find the subject matter engaging or stimulating. In contrast, ChatGPT, though proficient in creating coherent text, fundamentally lacks the authentic creativity and emotional nuances that children possess. Its responses, derived from algorithmic patterns in its training data, fail to capture the genuine imaginative spirit and enthusiasm that typically infuse children's exploratory narratives on science themes.

Thirdly, the disparity may emanate from the input constraints from the children and ChatGPT. ChatGPT is trained on a varied spectrum of internet texts and hence may exhibit more proficiency in generating text on themes more prevalently represented in its training data (Dwivedi et al., 2023). The predominance of nature-themed topics in the training data could account for ChatGPT's superior performance in nature-themed writing, showcasing a more affluent vocabulary and a higher utilization of idioms. To the opposite, although the nine-years-old children have likely been exposed to a more diverse set of subjects in their learning and everyday experiences, their vocabulary and language skills are still in a dynamic phase of development so that they may not exhibit the same level of vocabulary richness in nature-themed writing as ChatGPT.

In Tager-Flusberg's (2007) study, a positive correlation was identified between ToM skills and vocabulary-test scores in oral narratives. This suggests that individuals with advanced ToM abilities tend to have a strong command of communication-related verbs (e.g., John *said* that Mary is sleeping''), an issue relating to lexical complexity. In the current study, the lexical complexity was assessed through the complexity index that includes complex words, idioms, and unique terms. The results showed that in nature-themed Chinese writing, ChatGPT outperformed children in this complexity index, indicating a superior ToM performance regarding complexity in Chinese writing.

**4.4 Cohesion: children > ChatGPT**

The statistical showed that the nine-year-old children exhibited superior cohesion to ChatGPT in science-themed compositions. Another trend is that both groups displayed enhanced cohesion in nature-themed writing than in science-related writing.

Cohesion in writing is the use of linguistic techniques to connect ideas across sentences and paragraphs, thereby ensuring clarity and a logical flow of information. It involves mechanisms like pronouns, conjunctions, and transitions that seamlessly link parts of the text (Crossley & McNamara, 2012). In our experiment, ChatGPT infrequently employed causal connectives, such as "*所以*" and "*因此*" (meaning 'therefore'), averaging 5.23 connectives for each composition. Yet it manifests a propensity for the overuse of "*并/并且*" (meaning 'and'), a coordinating conjunction in Chinese, approximately three times in each composition. For instance, ChatGPT interconnected two clauses using "*并且*" in (4), rendering the whole sentence logically unusual in some sense. Furthermore, ChatGPT predominantly exhibits restricted utilization of conjunctive adverbs like "*即使*" (meaning 'even though') to forge cohesive links between sentences.

(4) "*房前的院子里有一只鹿在吃草，并且看起来非常宁静 (In the yard in front of the house, a deer is grazing, and it appears very peaceful)*"

Conversely, the children tend to leverage a more diversified array of cause-consequence connectives, such as "*于是*" ('therefore') and "*因此*" ('as a result/consequence'）, proficiently employing them to elucidate logical relations between clauses. Additionally, they used some colloquial connectives in writings, substituting "*结果*" for "*但*" (meaning 'but') as illustrated in example (5).

(5) "*我又偷偷的溜到它们后面偷鸭蛋，结果它们叫了一声 (I joyfully sneaked behind them to steal duck eggs, but they quacked loudly as a result).*"

Our findings corroborate those of Zhou et al. (2023), who highlighted that Chinese intermediate English majors outperformed ChatGPT in deep cohesion. These congruent findings likely stem from the inherent inability of text generation models including ChatGPT, as elucidated by Zhao et al. (2022). Present-day models struggle with cohesion more of then than not, leading to incoherent and poorly structured texts that hinder readers' comprehension. Despite progress in Natural Language Processing (NLP), the challenge still remain as how to improve textual cohesion. This issue arises from the statistical nature of these models, which often ignore a deep understanding of context and the meaning within the text.

ToM was also found to be positively correlated with local coherence in English narrative writing, which may stem from the necessity in English writing to understand social contexts, pragmatic weaknesses, and issues with perspective-taking (Brown & Klein, 2011). Local coherence refers to the clarity and logical connection between adjacent sentences or segments in a text, typically achieved through the use of connectives (Stede, 2022). In this study, the index of cohesion, which is evaluated by calculating the number of connectives, reflects local coherence mentioned by Brown and Klein (2011). That is, the nine-year-old children showed superior cohesion to ChatGPT in science-themed compositions in Chinese writing. This suggests that with respect to coherence related cohesion, nine-year-old children exhibit more abundant and delicate ToM than ChatGPT in Chinese narrative writing.

**4.5 Emotion: children > ChatGPT**

Our study revealed a pioneering finding that the children displayed a stronger intensity of emotion than ChatGPT in Chinese wring. A plausible explanation for this disparity is that emotions are often related to the subject or entity's vitality (TTaecharungroj & Stoica, 2023). Since ChatGPT has no entity, it lacks emotions inherently. It just fabricates responses by leveraging patterns discerned during its training phase. As a result, its outputs, though seemingly emotive, are devoid of genuine emotion but simply mirror human-like reactions, as substantiated by Zhao et al. (2022). In contrast, children possess authentic emotions, rendering their expressions more robust and palpable. This is exemplified vividly in science-themed compositions. For instance, a nine-year-old child employed a tag question to exude pride in their invention, as shown in (6). Despite the colloquialism, the sentences brim with childlike wonder and pronounced personal emotion.

(6) *"那这种书包甚至没有什么重量，假如你是一位大学生，那你的书包就跟小学生的书包差不多，怎么样，是不是特别神奇？"* (This schoolbag (backpack) isn't heavy at all. Imagine that you were a college student, then your schoolbag would be similar to that of an elementary school student. How about that? Isn't it quite amazing?).

Conversely, the passage generated by ChatGPT resembles emotionless tech broadcasts, as exemplified in (7). This stark contrast underscores ChatGPT's mechanical nature, devoid of human emotional nuances.

(7) *"智能环境监测器是我未来的创新发明，它将通过精确的传感技术和智能算法，帮助人们监测和改善室内环境质量，提高生活质量"* (The intelligent environmental monitor is my innovative invention for the future. It will employ precise sensing technology and intelligent algorithms to help people monitor and improve indoor environmental quality, thereby enhancing the quality of life).

However, our findings diverge from the study by Zhao et al. (2023), which attributed to ChatGPT the promising capacities for emotional response generation in dialogues. This inconsistency could stem from Zhao et al. (2023) juxtaposing the ChatGPT's performance of ChatGPT with other supervised baseline models' other than with real human-writers. Additionally, the focal point of Zhao et al. (2023) was dialogue, as opposed to our topic on Chinese essay narrative composition. Dialogues are typically concise, consisting of single sentences, while description requires multi-sentence logical coherence. Thus, synthesizing emotionally resonant description remains a challenge for models like ChatGPT.

According to Kosinski (2023), ChatGPT's ToM is on par with that of a nine-year-old child. In our study, ChatGPT showcased a lower intensity of emotional expression as compared to a nine-year-old child, suggesting a possibly lesser degree of ToM in the AI. This observation aligns with the concerns raised by Songchun Zhu (Zhou et al., 2023). Zhu questioned the effectiveness of traditional testing methods for evaluating machines' ToM development and their performing these tasks without actually possessing ToM. This brings into focus the complexity of measuring and understanding ToM in artificial intelligence, particularly in contexts that require nuanced emotional comprehension and expression.

## 5. Conclusion

This study probed into an intriguing question: how is ChatGPT comparable to nine-year-old children in Chinese writing? To address this issue, we undertook a comparative analysis of the Chinese compositions produced by ChatGPT and nine-year-old Chinese children, focusing on five indices to assess linguistic proficiency, fluency, accuracy, complexity, cohesion, and emotion. The results unveiled that the children performed better than ChatGPT in fluency, cohesion, and emotion whereas ChatGPT surpassed the children in accuracy of expression. As for complexity, the children exhibited superiority to ChatGPT in science-themed compositions, but the situation was reversed in nature-themed compositions.

Our study introduces an often-overlooked aspect "emotion" as an additional index to enrich the evaluative framework for Chinese writing proficiency and compares ChatGPT with nine-year-old children in Chinese writing proficiency from multiple perspectives. The results show that children exhibit more powerful emotional expressions than ChatGPT in Chinese writing, offering valuable insights for improving AI language models in their emotional understanding of texts. Additionally, our findings not only confirm Kosinski's (2023) claim that ChatGPT's ToM is equivalent to that of a nine-year-old child but also go beyond further, revealing that ChatGPT performed even worse than a nine-year-old child in the ToM regarding Chinese writing.

To conclude, this study is desired to provide inspiring insights into large language models and Chinese writing. For the former, ChatGPT is highly proficient in crafting accurate compositions but obviously deficient in cohesion and emotional expression, underscoring the imperative for future advancements of AI-driven writing, especially in forging sophisticated logical links and conveying emotion. For the latter, educators ought to motivate children of this age to refine the precision of their writing, focusing on aspects like punctuation and grammatical structures to enhance their comprehensive writing proficiency. The linguistic facets of fluency, accuracy, and complexity could serve as benchmarks to assess the efficacy of writing instruction in classrooms, thereby optimizing instructional design.

However, our research presents some constraints. First of all, this study was centered on Chinese compositions by ChatGPT and nine-year-old children, leaving native Chinese adults' writing unexplored.

Next, the investigation was confined to narrative compositions and didn't investigate other types of writings like argumentation, prose, and poems in Chinese. Subsequent exploration should broaden the genres of writing to showcase a full panorama by ChatGPT and human writers in Chinese writings.

Table 2. Five linguistic dimensions of Chinese writing from ChatGPT and 9-year-old children across nature and science-themed writing type

| | ChatGPT | | 9-year-old children | | | Group | Writing Type | Group x Writing Type |
|---|---|---|---|---|---|---|---|---|
| | Na-themed | Sci-themed | Na-themed | Sci-themed | | | | |
| Fluency | - .21 | -1.09 | .70 | .60 | *F-value* | 8.62 | 1.24 | .78 |
| | | | | | *p* | < .01 | .27 | .38 |
| | | | | | $\eta^2p$ | .07 | .01 | .01 |
| Accuracy | -1.62 | -2.73 | 1.90 | 2.45 | *F-value* | 226.15 | .92 | 8.30 |
| | | | | | *p* | < .001 | .34 | < .01 |
| | | | | | $\eta^2p$ | .66 | .01 | .07 |
| Complexity | 7.41 | -14.39 | 6.58 | .39 | *F-value* | 4.61 | 18.59 | 5.79 |
| | | | | | *p* | < .05 | < .001 | < .05 |
| | | | | | $\eta^2p$ | .04 | .14 | .05 |
| Cohesion | 3.01 | -4.16 | 2.44 | -1.30 | *F-value* | 4.11 | 92.53 | 9.17 |
| | | | | | *p* | < .05 | < .001 | < .01 |
| | | | | | $\eta^2p$ | .03 | .44 | .07 |
| Emotion | .05 | - .15 | .07 | .03 | *F-value* | 34.82 | 54.62 | 23.26 |
| | | | | | *p* | < .001 | < .001 | < .001 |
| | | | | | $\eta^2p$ | .23 | .32 | .17 |

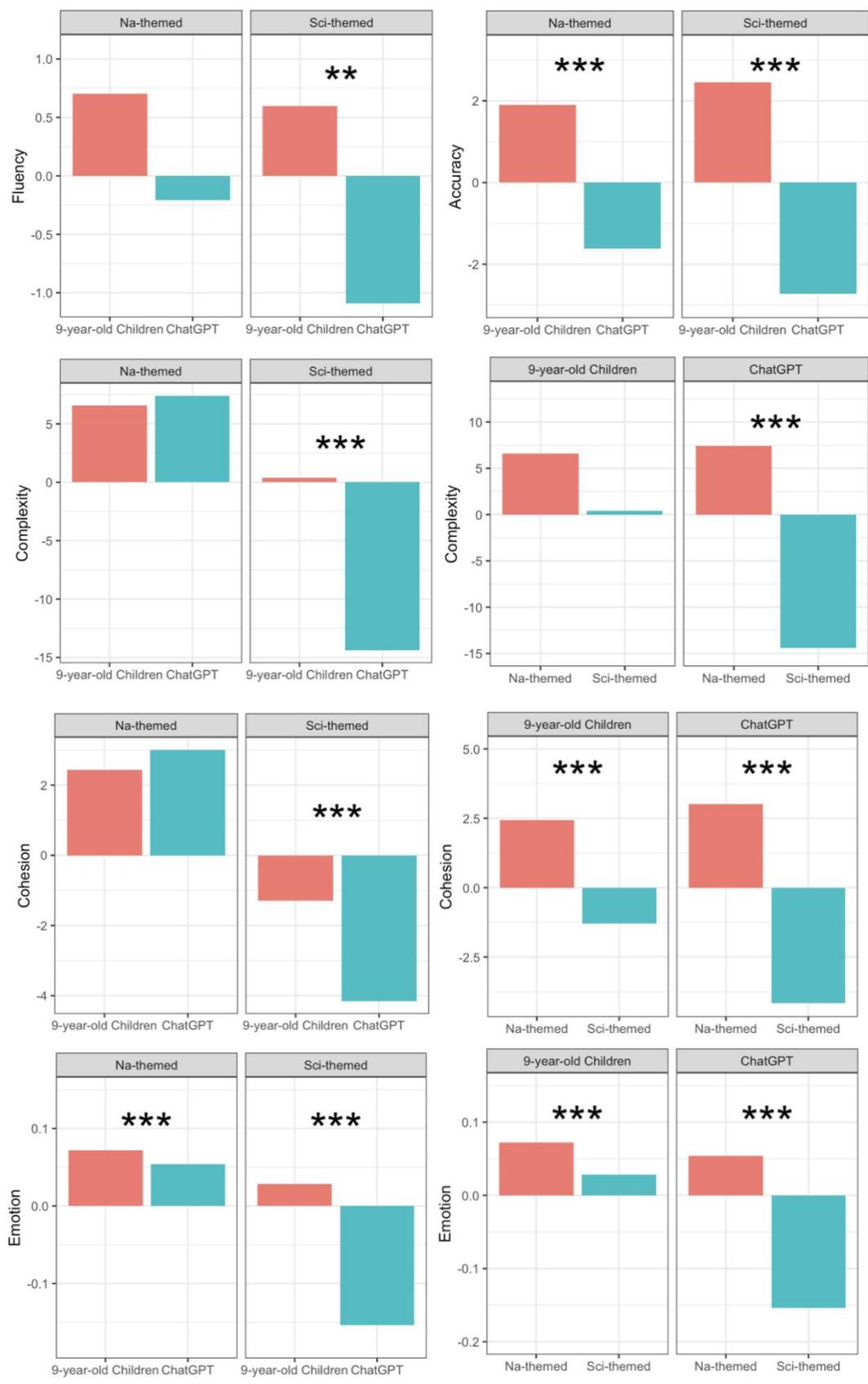

Figure 2. Five linguistic dimensions of Chinese writing produced by ChatGPT and 9-year-old children